\documentclass[11pt]{article}
\usepackage{eacl2017}
\usepackage{times}
\usepackage{url}
\usepackage{latexsym}
\usepackage{amsmath}
\usepackage{graphicx}
\usepackage{amssymb}
\eaclfinalcopy 

\title{Structural Attention Neural Networks for improved sentiment analysis}

\author{Filippos Kokkinos \\
  School of E.C.E., \\
  National Technical University of Athens,\\
  15773 Athens, Greece \\
  {\tt el11142@central.ntua.gr} \\\And
  Alexandros Potamianos \\
  School of E.C.E., \\
  National Technical University of Athens, \\
  15773 Athens, Greece \\
  {\tt potam@central.ntua.gr}
 }
\date{}

\begin{document}
\maketitle
\begin{abstract}
	We introduce a tree-structured attention neural network for sentences and small phrases and apply it to the problem of sentiment classification. Our model expands the current recursive models by incorporating structural information around a node of a syntactic tree using both  bottom-up and top-down information propagation. Also, the model utilizes structural attention to identify the most salient representations during the construction of the syntactic tree. To our knowledge, the proposed models achieve state of the art performance on the Stanford Sentiment Treebank dataset. 
\end{abstract}

\section{Introduction}
Sentiment analysis deals with the assessment of opinions, speculations, and emotions in text  \cite{Zhang:2012:DWS:2151163.2151168,Pang:2008:OMS:1454711.1454712}. It is a relatively recent research area that has attracted great interest as demonstrated by a series of shared evaluation tasks, e.g., analysis of tweets \cite{nakov2016semeval}. In \cite{turney2002unsupervised}, the affective ratings of unknown words were predicted utilizing the affective ratings of a small set of words (seeds) and the semantic relatedness between the unknown and the seed words. An example of sentence-level analysis was proposed in \cite{malandrakis2013sail}. Other application areas include the detection of public opinion and prediction of election results \cite{singhal2015modeling}, correlation of mood states and stock market indices \cite{bollen2011twitter}.

Recently, Recurrent Neural Network (RNN) with Long-Short Term Memory (LSTM) \cite{Hochreiter:1997:LSM:1246443.1246450} or Gated Recurrent Units (GRU) \cite{DBLP:journals/corr/ChungGCB14} have been applied to various Natural Language Processing tasks. Tree structured neural networks, which are found in literature as Recursive Neural Networks, hold a linguistic interest due to their close relation to syntactic structures of sentences being able to capture distributed information of structure such as logical terms\cite{Socher:2012:SCT:2390948.2391084}. These syntactic structures are N-ary trees which represent either the underlying structure of a sentence, known as constituency trees or the relations between words known as dependency trees.

This paper focuses on sentence-level sentiment classification of movie reviews using syntactic parse trees as input for the proposed networks. In order to solve the task of sentiment analysis of sentences, we work upon a variant of Recursive Neural Networks which recursively create representation following the syntactic structure. The proposed computation model exploits information from subnodes as well as parent nodes of the node under examination. This neural network is referred to as Bidirectional Recursive Network \cite{DBLP:journals/corr/IrsoyC13}. The model is further enhanced with memory units and the proposed structural attention mechanism. It is observed that different nodes of a tree structure hold information of variable saliency. Not all nodes of a tree are equally informative, so the proposed model selectively weights the contribution of each node regarding the sentence level representation using structural attention model.

We evaluate our approach on the sentence-level sentiment classification task using one standard movie review dataset \cite{socher2013recursive}. Experimental results show that the proposed model outperforms the state-of-the art methods.

\section {Tree-Structured GRUs}
Recursive GRUs (Tree-Gru) upon tree structures are an extension of the sequential GRUs that allow information to propagate through network topologies. 
Similar to Recursive LSTM network on tree structures \cite{DBLP:journals/corr/TaiSM15}, for every node of a tree, the Tree-GRU  has gating mechanisms that modulate the flow of information inside the unit without the need of a separate memory cell. The activation $h_j$ of Tree-GRU for node $j$ is the interpolation of the previous calculated activation $h_{jk}$ of its $kth$ child out of $N$ total children and the candidate activation ${\widetilde{h}}_j$ .
\begin{equation}
h_j = z_j * \sum_{k=1}^{N}h_{jk} + (1 - z_j)*{\widetilde{h}}_j
\end{equation}
where $z_j$ is the update function which decide the degree of update that will occur on the activation based on the input vector $x_j$ and previously calculated representation $h_{jk}$ :
\begin{equation}\label{eq:2}
z_j = \sigma(U_z*x_j + \sum_{k=1}^{N}W_z^i*h_{jk})
\end{equation}

The candidate activation ${\widetilde{h}}_j$ for a node $j$ is computed similarly to that of a Recursive Neural Network as in \cite{Socher:2011:SRA:2145432.2145450}:
\begin{equation}\label{eq:3}
{\widetilde{h}}_j = f(U_h*x_j + \sum_{k=1}^{N} W_h^k *(h_{jk} * r_j))
\end{equation}
where $r_j$ is the reset gate which allows the network to forget effectively previous computed representations when the value is close to 0 and it is computed as follows:
\begin{equation}\label{eq:4}
r_j = \sigma(U_r*x_j + \sum_{k=1}^{N}W_r^k*h_{jk})
\end{equation}

Every part of a gated recurrent unit $x_j,h_j,r_j,z_j,{\widetilde{h}}_j\in\mathbb{R}^d$ where d is the input vector dimensionality. $\sigma$ is the sigmoid function and $f$ is the non-linear tanh function.The set of matrices $W^k, U\in\mathbb{R}^{dxd}$ used in \ref{eq:2} - \ref{eq:4} are the trainable weight parameters which connect the $kth$ children node representation with the $jth$ node representation and the input vector $x_j$.

\subsection{Bidirectional TreeGRU}
A natural extension of Tree-Structure GRU is the addition of a bidirectional approach. TreeGRUs calculate an activation for node $j$ with the use of previously computed activations lying lower in the tree structure. The bidirectional approach for a  tree structure uses information both from under and lower nodes of the tree for a particular node $j$. In this manner, a newly calculated activation incorporates content from both the children and the parent of a particular node. \\
The bidirectional neural network can be trained in two seperate phases: i) the Upward phase and ii) the Downward phase. During the Upward phase, the network topology is similar to the topology of a TreeGru, every activation is calculated based on the previously calculated activations which are found lower on the structure in a bottom up fashion. When every activation has been computed, from leaves to root, then the root activation is used as input of the Downward phase. The Downward phase calculates the activations for every child of a node using content from the parent in a top down fashion. The process of computing the internal representations between the two phases is separated, so in a first pass the network compute the upward activation and after this is completed, then the downward representations are computed.\\
The upward activation $h^\uparrow_j$ similarly to TreeGRU for node $j$ is the interpolation of the previous calculated activation
$h^\uparrow_{jk}$ of its kth child out of N total children and the candidate activation ${\widetilde{h}}^\uparrow_j$. \\

\begin{equation}
h^\uparrow_j = z^\uparrow_j * \sum_{k=1}^{N}h^\uparrow_{jk} + (1 - z^\uparrow_j)*{\widetilde{h}}^\uparrow_j
\end{equation}

The update gate, rest gate and candidate activation are computed as follows:
\begin{equation}
z^\uparrow_j = \sigma(U_z*x^\uparrow_j + \sum_{k=1}^{N}W_z^k*h^\uparrow_{jk})
\end{equation}
\begin{equation}
r^\uparrow_j = \sigma(U_r*x^\uparrow_j + \sum_{k=1}^{N}W_r^k*h^\uparrow_{jk})
\end{equation}
\begin{equation}
{\widetilde{h}}^\uparrow_j = f(U_h*x^\uparrow_j + \sum_{k=1}^{N}W_r^k*(h^\uparrow_{jk} * r^\uparrow_j))
\end{equation}\\

\begin{figure}[!tbp]
\includegraphics[scale=0.42]{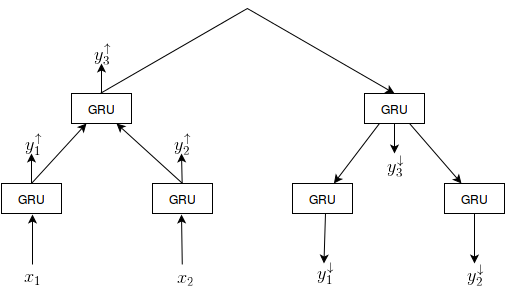}
\caption{A tree-structured bidirectional neural network with Gated Recurrent Units. The input vectors $x$ are given to the model in order to generate the phrase representations $y^\uparrow$ and $y^\downarrow$.}
\label{fig:figure1}
\end{figure}

The downward activation $h^\uparrow_j$ for node $j$ is the interpolation of the previous calculated activation $h^\uparrow_{p(j)}$, where the function $p$ calculates the index of the parent node, and the candidate activation ${\widetilde{h}}^\downarrow_j$.
\begin{equation}
h^\downarrow_j = z^\downarrow_j * h^\downarrow_{p(j)} + (1 - z^\downarrow_j)*{\widetilde{h}}^\downarrow_j
\end{equation}
The update gate, reset gate and candidate activation for the downward phase are computed as follows:
\begin{equation}
z^\downarrow_j = \sigma(U^d_z*h^\uparrow_j + W^d_z*h^\downarrow_{p(j)})
\end{equation}
\begin{equation}
r^\downarrow_j = \sigma(U^d_r*h^\uparrow_j + W^d_r*h^\downarrow_{p(j)})
\end{equation}
\begin{equation}
{\widetilde{h}}^\downarrow_j = f(U^d_h*h^\uparrow_j + W^d_h *(h^\downarrow_{p(j)} * r^\downarrow_j))
\end{equation}

During downward phase, matrix $U^d\in\mathbb{R}^{dxd}$ connects the upward representation of node $j$ with the respective $jth$ downward node while $W^d\in\mathbb{R}^{dxd}$ connect the parent representation $p(j)$.

\subsection {Structural Attention}
We introduce Structural Attention, a generalization of sequential attention model \cite{DBLP:journals/corr/LuongPM15} which extracts informative nodes out of a syntactic tree and aggregates the representation of those nodes in order to form the sentence vector. We feed representation $h_j$ of node through a one-layer Multilayer Perceptron with $W_w\in\mathbb{R}^{dxd}$ weight matrix to get the hidden representation $u_j$. 
\begin{equation}
u_j = tanh(W_w*h_j)
\end{equation}
Using the softmax function, the weights $a_j$ for each node are obtained based on the similarity of the hidden representation $u_j$ and a global context vetor $u_w\in\mathbb{R}^d$. The normalized weights $a_j$ are used to form the final sentence representation $s\in\mathbb{R}^d$ which is a weighted summation of every node representation $h_j$. 
\begin{equation}
a_j = \frac{u_j^\top * u_w}{\sum_{i=1}^{N}u_i^\top * u_w}
\end{equation}
\begin{equation}
s = \sum_{i=1}^{N} a_ih_i
\end{equation}
The proposed attention model is applied on structural content since all node representations contain syntactic structural information during training because of the recursive nature of the network topology.

\section{Experiments}
We evaluate the performance of the aforementioned models on the task of sentiment classification of sentences sampled from movie reviews. We use the Stanford Sentiment Treebank \cite{socher2013recursive} dataset which contains sentiment labels for every syntactically plausible phrase out of the 8544/1101/2210 train/dev/test sentences. Each phrase is labeled with respect to a 5-class sentiment value, i.e. very negative, negative, neutral, positive, very positive. The dataset can also be used for a binary classification subtask by excluding any neutral phrases for the original splits. The binary classification subtask is evaluated on  6920/872/1821 train/dev/test splits.

\subsection{Sentiment Classification}
For all of the aforementioned architectures at each node j we use a softmax classifier to predict the sentiment label $\hat{y}_j$. For example, the predicted label $\hat{y}_j$ corresponds to the sentiment class of the spanned phrase produced from node j. The classifier for unidirectional TreeGRU architectures uses the hidden state $h_j$ produced from recursive computations till node j using a set {$x_j$} of input nodes to predict the label as follows:
\begin{equation}
\hat{p}_{\theta}(y | {x_j}) = softmax(W_s*h_j)
\end{equation}
where $W_s\in\mathbb{R}^{dxc}$ and $c$ is the number of sentiment classes.

The classifier for bidirectional TreeBiGRU architectures uses both the hidden state $h^\uparrow_j$ and $h^\downarrow_j$ produced from recursive computations till node j during Upward and Downward Phase using a set {$x_j$} of input nodes to predict the label as follows:
\begin{equation}
\hat{p}_{\theta}(y | {x_j}) = softmax(W^\uparrow_s*h^\uparrow_j + W^\downarrow_s*h^\downarrow_j)
\end{equation}
where $W^\uparrow_s,W^\downarrow_s\in\mathbb{R}^{dxc}$ and $c$ is the number of sentiment classes. The predicted label $\hat{y}_j$ is the argument with the maximum confidence:
\begin{equation}
\hat{y}_j = \underset{y}{\mathrm{argmax}}(\hat{p}_{\theta}(y | {x_j}))
\end{equation}
For the Structural Attention models, we use for the final sentence representation $s$ to predict the sentiment label $\hat{y}_j$ where j is the corresponding root node of a sentence.
The cost function used is the negative log-likelihood of the ground-truth label $y^k$ at each node:
\begin{equation}
E(\theta) =  \sum_{k=1}^{m}\hat{p}_{\theta}(y^k | {x}^k) + \frac{\lambda}{2}||\theta||^2
\end{equation}
where m is the number of labels in a training sample and $\lambda$ is the L2 regularization hyperparameter. 
\begin{table}[!htbp]
\fontsize{11}{9}\selectfont
\centering
\label{my-label}
\begin{tabular}{lcc}
\textbf{Network Variant} & \multicolumn{1}{l}{d} & \multicolumn{1}{l}{$\left|\theta\right|$} \\ \hline
TreeGRU                 & \multicolumn{1}{l}{}           & \multicolumn{1}{l}{}                  \\
-without attention       & 300                            & 7323005                               \\
-with attention          & 300                            & 7413605                               \\
TreeBiGRU                &                                &                                       \\  
-without attention       & 300       & 8135405          \\  
-with attention          & 300                            & 8317810                               \\ \hline
\end{tabular}
\caption{Memory dimensions d and total network parameters $\left|\theta\right|$ for every network variant evaluated}
\end{table}
\subsection{Results}
The evaluation results are presented in Table 2 in terms of accuracy, for several state-of-the-art models proposed in the literature as well as for the TreeGRU and TreeBiGRU models proposed in this work.  Among the approaches reported in the literature, the highest accuracy is yielded by DRNN and DMN for the binary scheme (88.6), and by DMN for the fine-grained scheme (52.1). We observe that the best performance is achieved by TreeBiGRU with attention, for both binary (89.5) and fine-grained (52.4) evaluation metrics, exceeding any previously reported results. In addition, the attentional mechanism employed in the proposed TreeGRU and TreeBiGRU models improve the performance for both evaluation metrics. 

\begin{table}[!htbp]
\fontsize{11}{9}\selectfont
\centering
\label{dimensions}
\begin{tabular}{lcc}
\hline\hline
System           & \multicolumn{1}{l}{Binary} & \multicolumn{1}{l}{Fine-grained} \\
\hline
RNN           & 82.4                       & 43.2                             \\
MV-RNN           & 82.9                       & 44.4                             \\
RNTN           & 85.4                       & 45.7                             \\
PVec   & 87.8                       & 48.7                             \\
TreeLSTM         & 88.0                       & 51.0                             \\
DRNN           & 86.6                       & 49.8                             \\
DCNN            & 86.8                       & 48.5                             \\
CNN-multichannel & 88.1                       & 47.4                             \\
DMN            & 88.6              & 52.1                    \\
\hline
TreeGRU             &                            &                                  \\
- without attention & 88.6                          & 50.5                             \\
- with attention    & 89.0                          & 51.0                             \\
TreeBiGRU           &                            &                                  \\
- without attention & 88.5                          & 51.3                             \\
- with attention    & \textbf{89.5}                       & \textbf{52.4}  
\end{tabular}
\caption{Test Accuracies achieved on the Stanford Sentiment Treebank dataset. RNN, MV-RNN and RNTN \cite{socher2013recursive}. PVec: \cite{mikolov2013distributed}. TreeLSTM \cite{DBLP:journals/corr/TaiSM15}. DRNN \cite{DBLP:journals/corr/IrsoyC13}. DCNN \cite{DBLP:journals/corr/KalchbrennerGB14}.CNN-multichannel \cite{DBLP:journals/corr/Kim14f}. DMN \cite{DBLP:journals/corr/KumarISBEPOGS15}}  
\end{table}
\section{Hyperparameters and Training Details}
The evaluated models are trained using the AdaGrad \cite{Duchi:EECS-2010-24} algorithm using 0.01 learning rate and a minibatch of size 25 sentences. L2-regularization is performed on the model parameters with a $\lambda$ value $10^{-4}$. We use dropout with probability 0.5 on both the input layer and the softmax layer.\\
The word embeddings are initialized using the public available Glove vectors with a 300 dimensionality. The Glove vectors provide 95.5\% coverage for the SST dataset. All initialized word vectors are finetuned during the training process along with every other parameter. Every matrix is initialized with the identity matrix multiplied by 0.5 except for the matrices of the softmax layer and the attention layer which are randomly initialized from the normal Gaussian distribution. Every bias vectors is initialized with zeros. \\
The training process lasts for 40 epochs. During training, we evaluate the network 4 times every epoch and keep the parameters which give the best root accuracy on the development dataset.

\section {Conclusion}
In this short paper, we propose an extension of Recursive Neural Networks that incorporates a bidirectional approach with gated memory units as well as an attention model on structure level. The proposed models were evaluated on both fine-grained and binary sentiment classification tasks on a sentence level. Our results indicate that both the direction of the computation and the attention on a structural level can enhance the performance of neural networks on a sentiment analysis task.

\bibliography{eacl2017}
\bibliographystyle{eacl2017}

\end{document}